\documentclass[11pt,a4paper]{article}
\usepackage[hyperref]{eacl2021}
\usepackage{times}
\usepackage{latexsym}

\usepackage{microtype}

\aclfinalcopy 
\def\aclpaperid{1273}

\usepackage{helvet}  
\usepackage{courier}  
\usepackage{url}  
\usepackage{graphicx}  
\usepackage{latexsym}
\usepackage{subcaption}
\usepackage{url}
\usepackage{stmaryrd}
\usepackage{color}
\usepackage{xspace}
\usepackage{hyperref}       
\usepackage{url}            
\usepackage{booktabs}       
\usepackage{amsfonts}       
\usepackage{nicefrac}       
\usepackage{microtype}      
\usepackage{multirow}
\usepackage{amssymb}

\newcounter{notecounter}

\newcommand{\enotesoff}{\long\gdef\enote##1##2{}}
\newcommand{\enoteson}{\long\gdef\enote##1##2{{
\stepcounter{notecounter}
{\large\bf
\hspace{1cm}\arabic{notecounter} $<<<$ ##1: ##2
$>>>$\hspace{1cm}}}}}
\enoteson
\enotesoff

\def\balancedbert{17,748}
\def\balancedlstm{46,740}
\def\balancedbow{63,390}
\def\fbert{\,$\mathcal{F}_{\,\scriptsize\textsc{BERT}}$\,}
\def\fbow{\,$\mathcal{F}_{\,\scriptsize\textsc{BoW}}$\,}
\def\flstm{\,$\mathcal{F}_{\,\scriptsize\textsc{BiLSTM}}$\,}

\def\fhans{\,$\mathcal{F}_{\,\scriptsize\textsc{HANS}}$\,}

\def\xlnetbase{XLNET$_{\scriptsize \textrm{BASE}}$ }
\def\xlnetlarge{XLNET$_{\scriptsize \textrm{LARGE}}$ }
\def\bertbase{BERT$_{\scriptsize \textrm{BASE}}$ }
\def\bertlarge{BERT$_{\scriptsize \textrm{LARGE}}$ }
\def\ent{$E$\xspace}

\def\nent{$\neg E$\xspace}

\def\pph{$P$\xspace}
\def\npph{$\neg P$\xspace}
\def\nega{$N$\xspace}
\def\nnega{$\neg N$\xspace}

\newcommand{\bt}[2]{
	\multicolumn{1}{r}{\cellcolor{cyan!#1}{#2}}
}

\newcommand*{\affaddr}[1]{#1} 
\newcommand*{\affmark}[1][*]{\textsuperscript{#1}}

\title{Increasing Robustness to Spurious Correlations \\ using Forgettable Examples}

\author{\\
Yadollah Yaghoobzadeh{\rm\affmark[1]}~~ 
Soroush Mehri{\rm\affmark[2]}~~ 
Remi Tachet des Combes{\rm\affmark[2]}~~ \\
Timothy J. Hazen{\rm\affmark[1]}~~
Alessandro Sordoni{\rm\affmark[2]} \vspace{.15cm} \\
\affaddr{\affmark[1]Microsoft Turing, Montr\'eal}\\
\affaddr{\affmark[2]Microsoft Research, Montr\'eal}\\
\small \texttt{\{yayaghoo,alsordon\}@microsoft.com}
}

\begin{document}

\maketitle

\begin{abstract}
Neural NLP models tend to rely on spurious correlations between labels and input features to perform their tasks.
Minority examples, \textit{i.e.}, examples that contradict the spurious correlations present in the majority of data points, have been shown to increase the out-of-distribution generalization of pre-trained language models.
In this paper, we first propose using example forgetting to find minority examples without prior knowledge of the spurious correlations present in the dataset.
Forgettable
examples are instances either learned and then forgotten
during training or never learned.
We empirically show how these examples are related to minorities in our training sets.
Then, we introduce a new approach to robustify models by fine-tuning our models twice, first on the full training data and second on the minorities only.
We obtain substantial improvements in out-of-distribution generalization when applying our approach to the
MNLI, QQP, and FEVER datasets.

\end{abstract}

\section{Introduction}
Despite the impressive performance of current NLP models, these models often exploit spurious correlations: they tend to capture prediction correlations that hold for most examples but do not hold in general.
For instance, in natural language inference (NLI) datasets, word-overlap between hypothesis and premise is highly correlated with the \emph{entailment} label~\citep{linzen2019right,zhang-etal-2019-paws}. Therefore, these models are brittle when tested on examples that cannot be solved by recurring to these correlations, limiting their application in real-world scenarios. 
 \emph{Out-of-distribution} or \emph{challenging} sets are benchmarks carefully designed to break systems that rely on such correlations.


The paradigm of fine-tuning pre-trained language models (PLM) has pushed the state-of-the-art in a large variety of tasks involving natural language understanding (NLU)~\citep{devlin2018bert,wang2019superglue}.
This is achieved by self-supervised learning from an enormous amount of text.
PLMs also show increased robustness on challenging datasets \cite{Hendrycks19Using}.
This increase is attributed to an empirical finding that PLMs perform better on \emph{minority examples} present in the training data \cite{tu2020}. 
These minority examples violate the spurious correlations and therefore likely support the examples in challenging datasets.

\citet{tu2020} find minority examples by manually dividing the training data into two groups, according to the known spurious correlations (e.g., word-overlap in NLI). They present an analysis of the robustness of PLMs and its connection to minority examples. 
In this work, we first introduce a systematic way to find minority examples that does not need prior knowledge of spurious correlations, a big limitation of the earlier work. We then present a simple approach that increases the robustness of PLMs further by tuning models more on these examples.

To identify the set of minority examples, we adopt~\emph{example forgetting}~\citep{toneva2018empirical}. This statistic has been shown to relate to the hardness of examples, so we assume it is useful to find minorities in the training data.
Based on the definition presented in \citet{toneva2018empirical}, we consider an example \emph{forgettable} if during training it is
either properly classified at some point and misclassified later, or if it is never properly classified.
This method is model- and task-agnostic.
We show in our datasets that minority examples w.r.t to spurious correlations, such as word-overlap in NLI, are well represented in forgettable examples.

After finding minorities through forgettable examples, we propose a simple method to increase the robustness of PLMs further. We perform an additional fine-tuning on the minorities exclusively, after fine-tuning on the whole training data.
We find this strategy effective, as it increases robust accuracy, i.e., performance on out-of-distribution data, while minimally impacting performance on in-distribution examples. 
We evaluate our proposed methods in three tasks, including NLI  (MNLI,~\citealp{williams2017broad}), paraphrase identification (QQP,~\citealp{iyer2017first}) and fact verification (FEVER,~\citealp{thorne2018fever}). 
For each task, recent work has introduced out-of-distribution test sets targetting specific spurious correlations. 

Our contributions are the following:
\begin{itemize}
\item We propose using forgettable examples as a new approach for finding minority examples from training data without prior knowledge of spurious correlations. 
\item We show how to exploit minority examples and increase the robustness of deep neural models. 
This method outperforms other baselines in three challenging datasets: HANS~\citep{linzen2019right}, PAWS~\citep{zhang-etal-2019-paws} and FEVER-Symmetric~\citep{schuster2019towards}. 
Our method performs effectively when applied to both base and large versions of PLMs (e.g., \bertbase and \bertlarge).

\item We observe that finding minorities using a network shallower than the PLM is more effective to robustify it via fine-tuning.

\item 
We show that training models only on forgettable examples leads to poor performance in our datasets, which contrasts with the vision results from \citet{toneva2018empirical}. Our code is available at \url{github.com/sordonia/hans-forgetting}
\end{itemize}

\section{Datasets}
We consider three sentence pair classification tasks, namely natural language inference, paraphrase identification, and fact verification. 
In the following, we describe the datasets we choose for each task following an introduction of the task.
\label{sec:dataset}
\subsection{Natural Language Inference}
The first task we consider is MNLI~\citep{williams2017broad}, a common natural language inference dataset containing more than 400,000 premise and hypothesis pairs annotated with textual entailment information (\emph{neutral}, \emph{entailment} or \emph{contradiction}). Models trained on this dataset have been shown to capture spurious correlations, such as word-overlap between hypothesis and premise as a strong signal for the \emph{entailment} label~\citep{naik2018stress,linzen2019right}. 
A series of diagnostic out-of-distribution test sets have been devised to test robustness against such heuristics, e.g., HANS.

\textbf{HANS}~\citep[Heuristic Analysis for NLI Systems]{linzen2019right} is composed of both \emph{entailment} and \emph{contradiction} examples that have high word-overlap between hypothesis and premise (e.g. ``\emph{The president advised the doctor}'' $\longarrownot\longrightarrow$ ``\emph{The doctor advised the president}'').
A model relying exclusively on the word-overlap feature would not have a higher than chance classification accuracy on HANS. As a matter of fact, BERT~\citep{devlin2018bert} performance on this dataset is only slightly better than chance~\citep{linzen2019right}.
We consider HANS (size: 30k examples) and the MNLI matched dev \citep{williams2017broad} (size: 9815 examples) as our out- and in-distribution test sets for MNLI.



\subsection{Paraphrase Identification}
\label{sec:dataset_qqp}
QQP~\citep{iyer2017first} is a widely used dataset for paraphrase identification containing over 400,000 pairs of questions annotated as either paraphrase or non-paraphrase. 
As a consequence of the dataset design, pairs with high lexical overlap have a high probability of being paraphrases. Similarly to MNLI, models trained on QQP are thus prone to learning lexical overlap as a highly informative feature and do not capture the common sense underlying paraphrasing. PAWS dataset is designed to test that.

\textbf{PAWS}~\citep[Paraphrase Adversaries from Word Scrambling]{zhang-etal-2019-paws} is a question 
paraphrase dataset, 
well-balanced with respect to the lexical overlap heuristic. The accuracy of BERT is around 91.3\% on QQP and only 32.2\% on PAWS (Table \ref{tab:paws}). This makes it an interesting test-bed for our method. 
We use PAWS-QQP as our out-of-distribution set,
which contains 677 questions pairs. 
Training examples from PAWS were never used to update our models.
Following~\citet{zhang-etal-2019-paws} and ~\citet{utama2020mind}, our QQP training and testing splits are based on \citet{wang2017bilateral}.

\subsection{Fact Verification}
\label{sec:dataset_fever}
The task of fact verification aims to verify a claim given an evidence. The labels are \emph{support}, \emph{refutes}, and \emph{not enough information}. 
This task is defined as part of the 
Fact Extraction and Verification (FEVER) challenge~\citep{thorne2018fever}.
\citet{schuster2019towards} show that models ignoring evidence can still achieve high accuracy on FEVER.
They introduce an evaluation test set that challenges that bias.
Following~\citet{utama2020mind},
we use the \textbf{FEVER-Symmetric} datasets (Symm-v1 and Symm-v2 with 717 and 712 examples, respectively) for out-of-distribution evaluation\footnote{https://github.com/TalSchuster/FeverSymmetric}.

\section{Finding Minorities with Forgettables}
We first define example forgetting and how to compute it.
We then show that it can be used to find minority examples in the training data. 

\subsection{Forgettable examples} An example is forgotten if it goes from being correctly to incorrectly classified during training (each such occurrence is called a \emph{forgetting event}). This happens due to the stochastic nature of gradient descent, in which gradient updates performed on certain examples can hurt performance on others. If an example is forgotten at least once or is never learned during training it is dubbed~\emph{forgettable}. 
Finding forgettable examples entails training the model on $\mathcal{D}$ and tracking the accuracy of each example at each presentation during training. The algorithm for computing forgettability is cheap~\citep{toneva2018empirical} and only requires storing the accuracy of each particular example at each epoch.

In \citet{toneva2018empirical}, they extracted forgettable examples from a \emph{shallower network} compared to their target model. This makes finding forgettables more efficient and also results in a more diverse set of examples, as 
the number of forgettable examples is usually higher for weaker models. 
Another factor is that the shallow models exhibit less memorization due to their fewer number of hyperparameters \cite{sagawa2020investigation} and therefore their forgettables are potentially more representative of the minorities.

We compute forgettable examples using two models with  significantly lower capacity compared to PLMs. 
The first one is a ``siamese'' BoW classifier in which hypothesis and premise are independently encoded as a mean of word embeddings. This common model in NLP tasks has surprisingly good performance while relying only on the bag of lexical features. We also consider a siamese BiLSTM model.
More details can be found in Appendix \ref{sec:biased_models}.
Finally, for comparison, we also experiment with the
model used for HANS in SOTA baselines \cite{clark2019dont,utama2020mind} (see also \S\ref{sec:baselines}), as well as \bertbase for in NLI.



\begin{table}[t]
\centering
\begin{tabular}{llcc}
\toprule
\textbf{Dataset} & \textbf{Model} & $|\mathcal{F}|$ & \textbf{Dev Acc.} \\
\midrule
& BoW         & \balancedbow & 64.0\\
MNLI & BiLSTM & \balancedlstm  & 69.6\\
(392,703) & BERT       & \balancedbert & 84.5\\
\midrule
& BoW         &71,116  & 81.1\\
QQP 
& BiLSTM    &76,634   & 84.3 \\
(384,348) & BERT        &20,498   & 91.3 \\
\midrule
& BoW         & 76,368  & 53.3\\
FEVER
& BiLSTM      & 68,406  & 56.7 \\
(242,911)
& BERT        & 21,066   & 84.4 \\

\bottomrule
\end{tabular}
\caption{Number of ``forgettable'' examples along with the accuracy on the MNLI matched, QQP, and FEVER development set. BERT's forgettables are used only 
in MNLI experiments.
The full training size is shown in parenthesis for each dataset.}
\label{tab:forg_stats}
\end{table}

We train the shallow models for five epochs and track forgetting statistics after each epoch. Table~\ref{tab:forg_stats} shows the number of forgettable examples for BoW, BiLSTM and \bertbase on the MNLI, QQP and FEVER training sets. The performance of the models on the dev set of MNLI is also included. 

\begin{figure}[t]
\centering
  \includegraphics[width=0.48\textwidth]{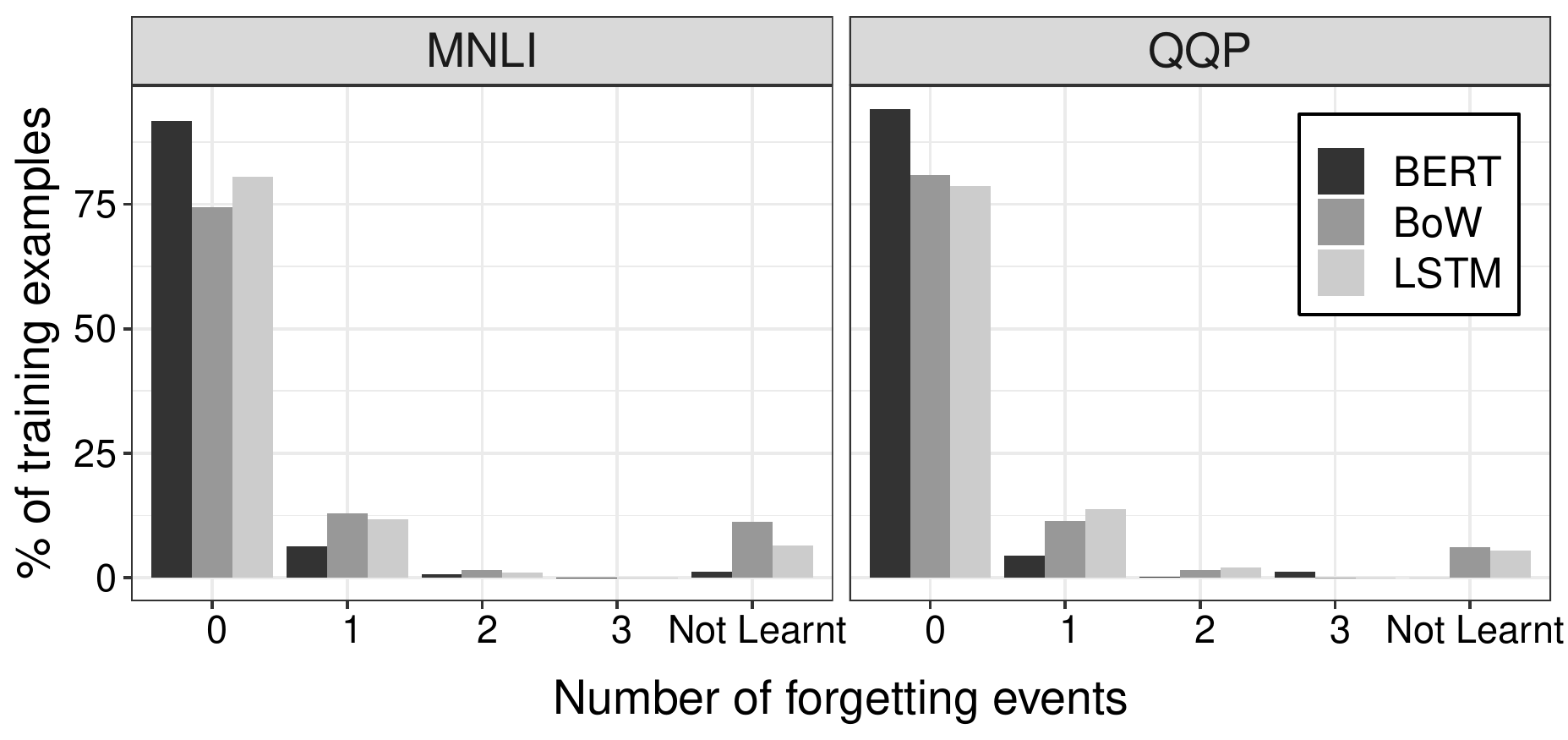}
  \caption{
  Forgetting events in MNLI and QQP training sets for the three models. A majority of examples are not forgotten during training. 
  }
\label{fig:forgcount-freq}
\end{figure}

The distribution of forgetting events for each model can be found in Fig~\ref{fig:forgcount-freq}. We see that a majority of examples are not forgotten. Most of the forgettables are either those that are never learned or have only one forgetting event. In what follows, we denote the sets of forgettable examples from \bertbase, BiLSTM, and BoW as \fbert, \flstm and \fbow respectively.

\subsection{Forgettable and minority examples}
\label{sec:forg-mionority}
We focus on two important spurious correlations: word-overlap and  contradiction-word. These correlations or biases are addressed in related work for MNLI and QQP \cite{tu2020,Zhou2020towards}.
For convenience, we use ``positive'' for either entailment, supports or paraphrase, and ``negative'' for contradiction, refutes or non-paraphrase

High word-overlap between two sentences is spuriously correlated to the positive label in all three datasets. 
In Table~\ref{tab:wordoverlap-labels}, we show that on average, positive examples have higher word-overlap compared to non-positive ones. 
In other words, minorities w.r.t. word-overlap correspond to non-positive examples with high word-overlap and positive examples with low word-overlap.
For MNLI and QQP, the distribution in \fbow and \flstm exhibit an interesting behavior: on average, non-positive examples have higher word-overlap. \fbert has the same average for both labels.
For FEVER, the difference is also clear as the gap in word-overlap between positive and non-positive examples is lower for forgettables.
The table allows to conclude that forgettable examples contain more minority examples than a random subset of the same size.

In Table~\ref{tab:negation-labels}, we perform a similar analysis for the presence of contradiction words in the second sentence, which is shown to correlate with negative class in MNLI  \cite{naik2018stress,Zhou2020towards} and FEVER \cite{schuster2019towards}. We choose these contradiction words: \{``not'', ``no'', ``doesn't'', ``don't'', ``never'', ``any''\}, and analyze all three datasets.
We observe here as well that forgettables contain more minority examples, as their percentage of examples with a contradiction word is lower for negative examples, which is the opposite than in the overall dataset (with the exception of \fbert and FEVER).

\begin{table}[t]
\small
    \centering
    \begin{tabular}{lc c | c c | c c  }
    \toprule
    & \multicolumn{2}{c}{MNLI} & \multicolumn{2}{c}{QQP} & \multicolumn{2}{c}{FEVER}\\
             &  \pph & \npph & \pph & \npph & \pph & \npph \\
    \midrule
         all & \textbf{.33} &  .22 & \textbf{.51} & .34
         & \textbf{.19} & .15\\
         \fbow & .26 &  \textbf{.30} & .48 & .49
         & .18 & .17\\
         \flstm & .25 & .28 & .48 & \textbf{.50}
         & .18 & 17\\
         \fbert & .26 &  .26 & .50 & .50
         & \textbf{.19} & \textbf{.18}\\
    \bottomrule
    \end{tabular}
    \caption{Average Jaccard index as a measure of word-overlap between two sentences grouped by \pph (positive) and \npph (non-positive).}
    \label{tab:wordoverlap-labels}
\end{table}

\section{Robustifying by Fine-Tuning on Minority Examples}
\begin{figure}[tbp]
\centering
\includegraphics[scale=0.55]{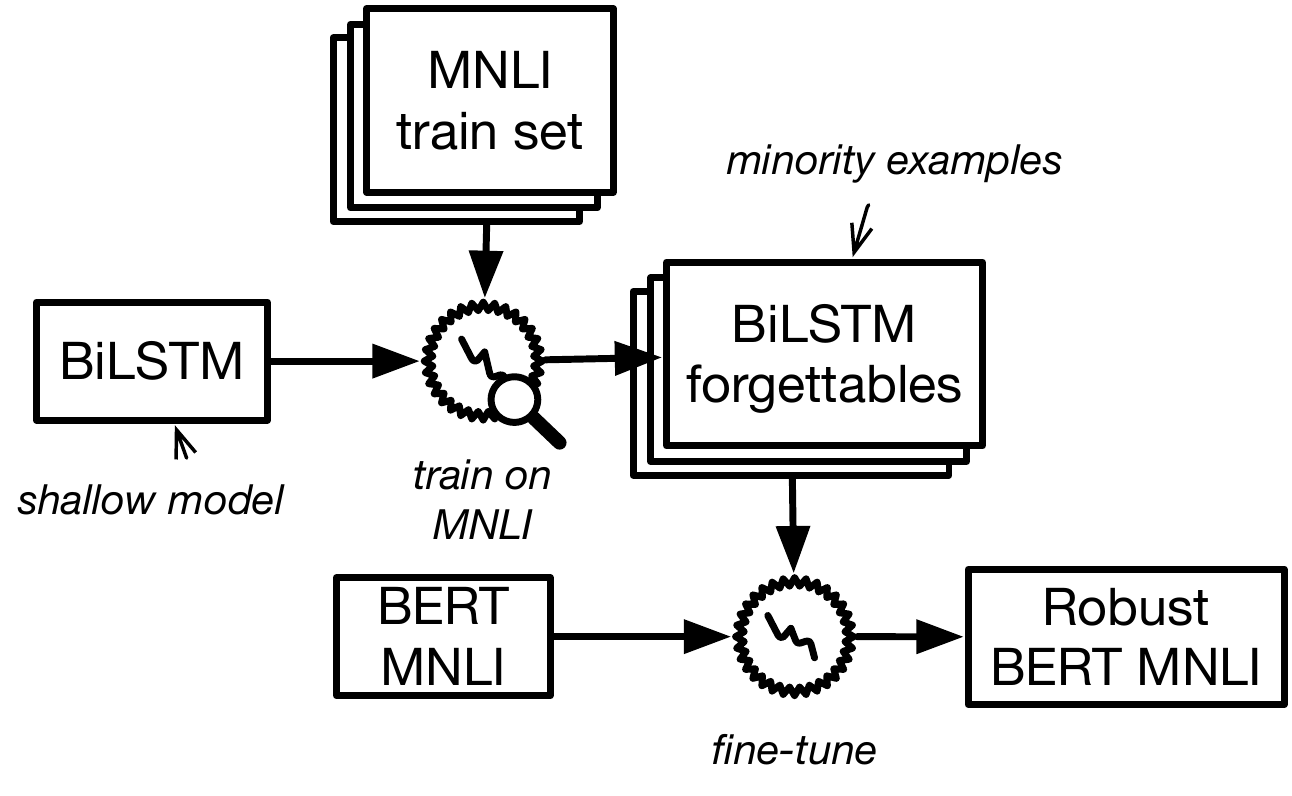}
\caption{Our proposed framework to train more robust models for an example dataset (MNLI).
We first detect minority examples through ~\emph{forgettable} examples of a shallow model from the training set. We then fine-tune a PLM model (e.g. BERT) in two rounds: first on the full training set, and second on the forgettable subset exclusively. The final model is more robust.
}
\label{fig:method}
\end{figure}

Prior work shows that PLMs generalize to out-of-distribution data because they generalize better on minority examples
from the training set \cite{tu2020}. Here, we introduce a simple approach that exploits minority examples to increase robustness.
In this approach, we fine-tune a PLM in two successive phases, first on the full training set, and then on the minority examples only.
Our method does not need changes to the training objectives.
An illustration is shown in Fig~\ref{fig:method}.



\subsection{PLMs}
\label{sec:strong}
We are interested in the robustness of large PLMs. In this work, we focus on two such models, BERT and XLNet, and experiment with both their base and large versions. \bertbase being the model of choice in previous work \citep{clark2019dont,zhang-etal-2019-paws,utama2020mind}, it will serve as our default architecture. We adopt the Transformers library \citep{Wolf2019HuggingFacesTS}. Our robust models are obtained by fine-tuning PLMs on the full training set for $3$ epochs (using the default hyperparameters for each task) and then on the forgettable examples only, for $3$ more epochs with a smaller learning rate.
See \ref{sec:hyperparameters_trainingtime} in Appendix for more details.

\begin{table}[t]
\small
\centering
\begin{tabular}{lc c | c c | cc}
\toprule
& \multicolumn{2}{c}{MNLI} & \multicolumn{2}{c}{QQP} & \multicolumn{2}{c}{FEVER} \\
&  \nega & \nnega  & \nega & \nnega & \nega & \nnega \\
\midrule
all & \textbf{31.5} &  10.4 & \textbf{6.2} & 4.0 & 
\textbf{16.4} & 1.2\\
\fbow & 9.9 &  11.5  & 4.1 & \textbf{4.6} & 1.6 & \textbf{3.1}\\
\flstm & 11.5  & 11.2 & 4.2 & 4.4 & 2.9 & 2.7\\
\fbert & 14.2 &  \textbf{12.3} & 4.2 & 4.4 & 6.3 & 2.5\\
\bottomrule
\end{tabular}
\caption{Percentage of examples containing one of the negative keywords in the hypothesis / second question / claim in MNLI / QQP / FEVER. We group examples by binary labels (\nega: negative and \nnega: non-negative) to show the distribution difference between forgettable and overall training examples.}
\label{tab:negation-labels}
\end{table}

\subsection{Baselines}
\label{sec:baselines}
Recently, multiple methods have been proposed to learn more robust models through mitigating biases~\citep{clark2019dont,he2019unlearn,mahabadi2019simple,utama2020mind}.
In these works, PLMs are fine-tuned on a re-weighted version of the source dataset, in which examples are weighted based on their hardness. Hardness is measured by training biased models using prior knowledge of the biases or spurious correlations, e.g., a linear model with word-overlap features for NLI.
Compared to these works, our method does not need prior knowledge of spurious correlations and exploit the minority examples explicitly by further fine-tuning on them.

We consider the recent \emph{confidence regularization} or ``Reg-conf'' technique from~\citet{utama2020mind}, as our main 
baseline in all three tasks. 
This method is an improvement to the earlier related work 
in making more robust NLP models \citep{he2019unlearn,clark2019dont}. Specifically, Reg-conf claims to maintain the in-distribution performance
while improving out-of-distribution and is doing so without introducing new hyperparamters.
For MNLI, we report the results of three other baselines of
\citet{he2019unlearn}, \citet{clark2018semi} and \citet{mahabadi2019simple}.

Previous work generally design biased models assuming a priori knowledge of the specific dataset biases~(with the exception of~\citet{he2019unlearn}, that use a BoW model for HANS). For HANS and PAWS,~\citet{clark2019dont} and~\citet{utama2020mind} employ a model with 7 input features, such as word-overlap between premise and hypothesis. To highlight the generality of our approach, we also add this biased model to the set of our shallow models for HANS and
fine-tune on its forgettables~(\fhans with the size of around 200k).
For FEVER-symmetric,~\citet{utama2020mind} consider an LSTM model that takes only the ``claim'' as input and ignores the ``evidence''. These baselines re-weight or confidence-regularize training examples using the biased models' performance.

\subsection{Results}
\label{sec:eval}
\subsubsection{MNLI and HANS}
\begin{table}[t]
\setlength{\tabcolsep}{2.pt}
\centering
\begin{tabular}{lccc}
\toprule
\textbf{Model} & \textbf{MNLI} & \textbf{HANS}  & \textbf{Avg.}  \\
\midrule
BERT & \textbf{84.4}$_{\pm 0.1}$ & 62.9$_{\pm 1.5}$ & 73.7$_{\pm 0.8}$\\
BERT+\fbert   
& 83.0$_{\pm 0.4}$ & 68.9$_{\pm 1.4}$ & 75.9$_{\pm 0.7}$  \\
BERT+\flstm 
& 82.9$_{\pm 0.4}$ & 70.4$_{\pm 0.9}$ & 76.7$_{\pm 0.5}$ \\
BERT+\fbow & 
 83.1$_{\pm 0.3}$  & \textbf{70.5}$_{\pm 0.7}$ & \textbf{76.8}$_{\pm 0.4}$ \\
\hspace{0.1cm} BERT + Rand$_{\balancedbow}$       
&  84.3                        & 63.6                     & 73.9  \\
BERT+\fhans & 
83.9$_{\pm 0.4}$ &	69.5$_{\pm 0.9}$ & 76.7$_{\pm 0.5}$ \\
\midrule
\midrule
\emph{\citet{clark2019dont}} & & & \\
Reweight & 83.5 & 69.2 & 76.4 \\
Learned Mixin & 84.3 & 64.0 & 74.2\\
\midrule
\multicolumn{3}{l}{\emph{\citet{mahabadi2019simple}}}
 &   \\
Product of Experts &  84.0 & 66.5 & 75.3     \\
\midrule
\emph{\citet{he2019unlearn}} & & &  \\
DRiFt-HYPO & 84.3 & 67.1 & 75.7     \\
\midrule
\emph{\citet{utama2020mind}} & & &  \\
Reg-conf$_{\mbox{\tiny{hans}}}$ & 
84.3$_{\pm 0.1}$ & 69.1$_{\pm 1.2}$ & 76.7$_{\pm 0.6}$     \\
\bottomrule
\end{tabular}
\caption{Results of our BERT models fine-tuned on different sources of forgettable examples. 
For each line, the accuracy on MNLI and HANS are shown, as well as their average. 
}
\label{tab:maintable}
\end{table}

In Table~\ref{tab:maintable}, we present the results of our models and four recent baselines.
The first line reports the performance of BERT on MNLI and HANS. The following lines report the results obtained by fine-tuning BERT on the set of forgettable examples obtained using different shallow models. We also report the average performance between MNLI and HANS. The results confirm that tuning the model towards minority examples improves robustness with a slight drop in MNLI accuracy.
Our best model is obtained by fine-tuning on \fbow, achieving a HANS mean accuracy of 70.5\% (with a max of 71.3\% over five seeds, which constitutes a +8.4\% absolute improvement w.r.t to the initial BERT). To assess whether \fbow is indeed responsible for the improvement, we also fine-tune BERT on the same number of randomly chosen examples (BERT + Rand$_{\balancedbow}$), which leads to a negligible improvement.

Fine-tuning on \flstm is comparable to fine-tuning on \fbow, which demonstrates that both BoW and BiLSTM models learn similar spurious correlations.
We also added results of fine-tuning BERT on its own forgettables for this task. Note that while it provides less improvement in robustness than on \flstm or \fbow\footnote{To eliminate the forgettables' size factor and focus on the type of model instead, we run an experiment where we sample from \fbow the same numbers as \fbert. The result of our fine-tuning on that smaller \fbow was still significantly better than \fbert.}, it does generate a significant 6.0\% increase in performance.
Finally, we also report fine-tuning results on \fhans, 
the biased model designed for HANS, and observe that it performs well with a smaller loss on MNLI and a smaller gain on HANS compared to \fbow and \flstm.

\begin{figure}[t]
\includegraphics[scale=0.40]{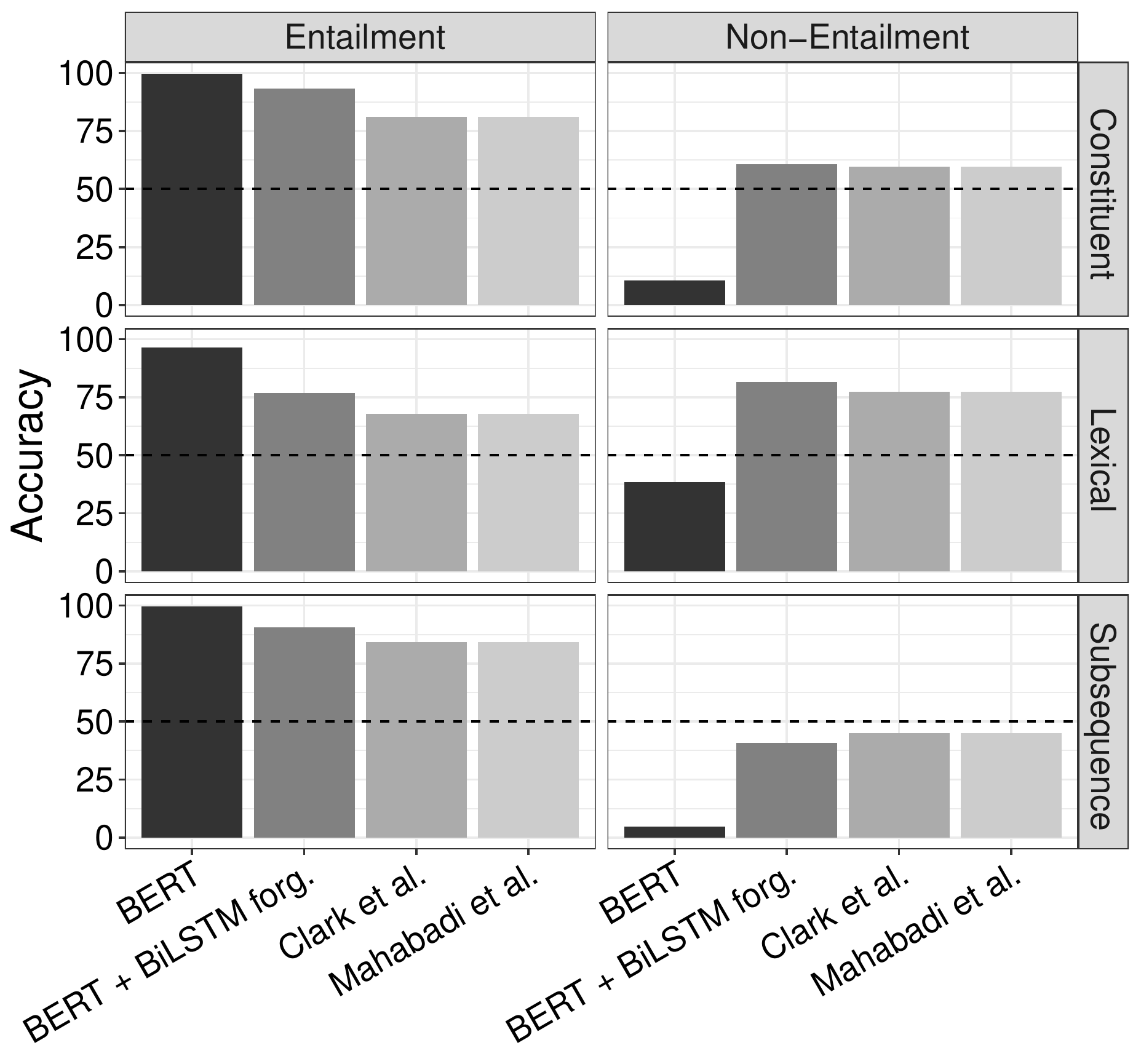}
\caption{Performance of our BERT fine-tuned on the BiLSTM forgettables \flstm, and baselines on the ``entailment'' and ``non-entailment'' categories for each heuristic HANS was designed to capture. 
}
\label{fig:fine_eval_baselines}
\end{figure}

Compared to other baselines, our approach achieves a comparable or better average accuracy of MNLI and HANS, despite its simplicity.
In Fig.~\ref{fig:fine_eval_baselines}, we breakdown the results of our best performing model for the three different heuristics HANS was built upon. Our method does not suffer as much as other baselines in the entailment class, and still provides a significant improvement for non-entailment.
(More analysis is presented in Appendix.)

\subsubsection{QQP and PAWS}
\label{sec:paws}

\begin{table}[t]
\setlength{\tabcolsep}{3.pt}
\centering
\begin{tabular}{lccc}
\toprule
\textbf{Model}                           & \multicolumn{1}{c}{\textbf{QQP}} & \multicolumn{1}{c}{\textbf{PAWS}} &           \multicolumn{1}{c}{\textbf{Avg.}}         \\
\midrule
BERT & \textbf{90.9$_{\pm 0.4}$} & 34.5$_{\pm 1.5}$ & 62.7$_{\pm 0.6}$ \\
BERT+\fbow   & 89.0$_{\pm 0.9}$ & \textbf{48.8$_{\pm 5.2}$} & \textbf{68.9$_{\pm 2.2}$}  \\
BERT+\flstm & 88.0$_{\pm 0.8}$ & 47.6$_{\pm 4.1}$  & 67.8$_{\pm 1.7}$ \\
\midrule
\midrule
\emph{\citet{utama2020mind}}\\
 BERT & 91.0 & 34.3 & 62.6 \\
 Reg-conf$_{\mbox{\small{hans}}}$ & 89.1  & 39.8& 64.5 \\
\bottomrule
\end{tabular}
\caption{Results of \bertbase trained on different sets of training examples. Accuracy (\%) is reported on the QQP test set (size: 10k) and the PAWS dev set (size: 677), alongside their average.
}
\label{tab:paws}
\end{table}

Here we report the results of our method applied to QQP and PAWS as out-of-distribution dataset. Results can be found in Table~\ref{tab:paws}. 
We observe that our method improves out-of-distribution accuracy substantially.
It is worth noting that the ground-truth labels in QQP contain noisy annotations~\citep{iyer2017first}; a portion of performance loss on QQP could be attributed to that.

Our method outperforms Reg-conf$_{\mbox{\small{hans}}}$, while being simpler in terms of both the biased model and the training regime.
We notice that Reg-conf$_{\mbox{\small{hans}}}$ also loses in-distribution performance\footnote{The authors report accuracy on each label individually and not the overall accuracy. We compute that based on their reported numbers.}.

\subsubsection{FEVER}
\label{sec:fever}
In Table~\ref{tab:fever}, we report the results of our method applied to the FEVER development and symmetric evaluation sets~(see \S\ref{sec:dataset_fever}). 
Our approach again works well for both \fbow and \flstm, but here we also gain on the original dev set when compared to the initial \bertbase results.
The gains of our method are larger than those of the Reg-conf$_{\mbox{\tiny{claim}}}$ baseline, which uses a biased model
tailored to FEVER-symmetric.

\begin{table}[t]
\setlength{\tabcolsep}{1.8pt}
\centering
\begin{tabular}{lcccc}
\toprule
\textbf{Model} & \multicolumn{1}{c}{\textbf{FEVER}} & \multicolumn{1}{c}{\textbf{Sym-v1}} &
\multicolumn{1}{c}{\textbf{Sym-v2}} \\
\midrule
BERT & 86.1$_{\pm 0.3}$ & 57.7$_{\pm 1.3}$ & 64.7$_{\pm 1.1}$ \\
BERT+\fbow & \textbf{87.1}$_{\pm 0.2}$ & 61.0$_{\pm 1.4}$ & \textbf{67.0}$_{\pm 1.5}$\\
BERT+\flstm    & 86.5$_{\pm 0.4}$ & \textbf{61.7}$_{\pm 1.2}$ & 66.6$_{\pm 1.0}$\\
\midrule
\midrule
\citet{utama2020mind} \\
BERT & 85.8$_{\pm 0.1}$ & 57.9$_{\pm 1.1}$ & 64.4$_{\pm 0.6}$\\
Reweighting$_{\mbox{\tiny{bigrams}}}$  & 85.5$_{\pm 0.3}$ & \textbf{61.7}$_{\pm 1.1}$ & 66.5$_{\pm 1.3}$\\
Reg-conf$_{\mbox{\tiny{claim}}}$ &  86.4$_{\pm 0.2}$ & 60.5$_{\pm 0.4}$& 66.2$_{\pm 0.6}$\\
\bottomrule
\end{tabular}
\caption{Accuracy of different FEVER trained models on
FEVER dev, and symmetric v1 and v2 datasets.}
\label{tab:fever}
\end{table}

\subsection{Analysis}

\begin{figure}
\centering
\includegraphics[scale=0.4]{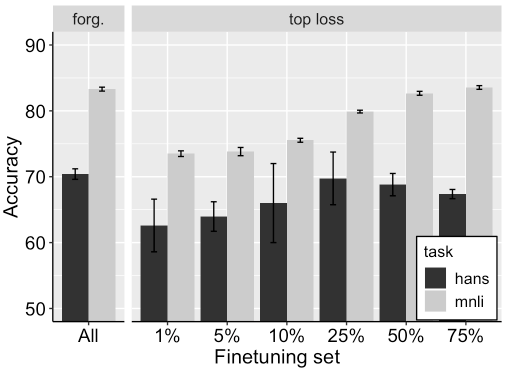}
\caption{Accuracy on MNLI and HANS when the fine-tuning set is picked
from examples in forgettables or in a range of percentages of examples with highest loss.}
\label{fig:loss_forg}
\end{figure}

\paragraph{Final loss to detect minority examples}
An alternative way to find examples from the minority is
to simply rank training examples based on their final loss value.
In Fig~\ref{fig:loss_forg}, we compare that with our method based on forgettables. The two are obviously related, as the examples that are never learned rank the highest w.r.t to the loss and are considered as forgettables. However, Fig~\ref{fig:loss_forg} shows, for MNLI and HANS, that using forgettables produces better performance both in- and out-of-distribution. 
One additional issue with using the final loss to pick examples is the need to determine either a threshold value $\alpha$ on the loss (keep examples with a loss larger than  $\alpha$) or a number $N$ of examples to retain.
The optimal $\alpha$ or $N$ might yield better performance but
finding them implies using the out-of-distribution set.

\begin{table}[t]
\centering
\begin{tabular}{lcc}
\toprule
\textbf{Train examples} &  \textbf{MNLI} & \textbf{HANS} \\
\midrule
\fbert $_{(\balancedbert)}$   & 49.6$_{\pm 0.2}$                     & 37.9$_{\pm 1.3}$\\
\hspace{0.2cm} Random $_{(\balancedbert)}$ 
                   & 74.7$_{\pm 0.4}$ & 50.8$_{\pm 0.2}$\\
\flstm $_{(\balancedlstm)}$   & 66.7$_{\pm 0.9}$ & 54.0$_{\pm 0.6}$                   \\
\hspace{0.2cm} Random $_{(\balancedlstm)}$                & 78.8$_{\pm 0.4}$ & 51.3$_{\pm 0.6}$    \\
\fbow $_{(\balancedbow)}$       & 68.2$_{\pm 0.8}$ & 55.4$_{\pm 1.4}$                  \\
\hspace{0.2cm} Random $_{(\balancedbow)}$             & 79.9$_{\pm 0.4}$ & 51.8$_{\pm 0.2}$     \\
\midrule
All & {84.5}$_{\pm 0.1}$  & 63.1$_{\pm 1.2}$ \\
\bottomrule
\end{tabular}
\caption{Results of \bertbase models fine-tuned on the set of forgettable examples only.}
\label{tab:forg_tuning_mnli}
\end{table}

\begin{figure}[h]
\centering
\includegraphics[scale=0.4]{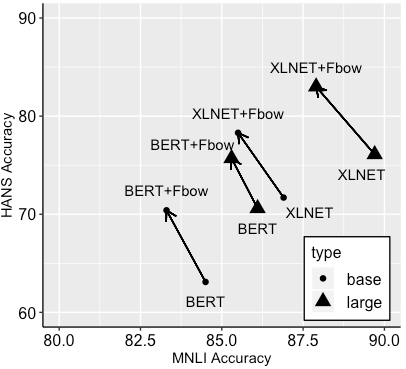}
\caption{MNLI vs HANS accuracy for both base and large versions of BERT and XLNET.}
\label{fig:base_large}
\end{figure}

\begin{table}[t]
\centering
\begin{tabular}{lll}
\toprule
\textbf{Model}         & \textbf{QQP}  & \textbf{PAWS} \\
\toprule
\bertbase         & 90.9 & 34.5 \\
\bertbase+\fbow   & 89.0 & 48.8 \\
\bertlarge        & \textbf{91.2} & 36.0 \\
\bertlarge+\fbow  & 88.3 & 54.4 \\
\midrule
\xlnetbase        & 90.9 & 37.1 \\
\xlnetbase+\fbow  & 88.2 & 55.2 \\
\xlnetlarge       & 89.2 & 48.4 \\
\xlnetlarge+\fbow & 87.8 & \textbf{65.2}\\
\bottomrule
\end{tabular}
\caption{Average accuracy across random seeds on QQP and PAWS
for BERT and XLNET base and large models, before and after
fine-tuning on \fbow.}
\label{tab:xlnet_qqp}
\end{table}

\begin{table}[]
\centering
\begin{tabular}{lcc}
\toprule
\textbf{Model}                                               & \textbf{FEVER} & \textbf{Symm-v1} \\
\toprule
\bertbase                       & 86.1  & 57.7          \\
\bertbase+\fbow  & 87.2  & 61.0          \\
\bertlarge                           & 86.9  & 59.7          \\
\bertlarge+\fbow      & 86.5  & 67.8          \\
\midrule
\xlnetbase                      & 86.4  & 63.9          \\
\xlnetbase+\fbow & 87.5  & 67.8          \\
\xlnetlarge                          & 88.2  & 68.8          \\
\xlnetlarge+\fbow     & \textbf{88.7}  & \textbf{75.3}         \\
\bottomrule
\end{tabular}
\caption{Average accuracy over seeds on FEVER and Fever-symm-v1
for BERT and XLNET base and large models, before and after
fine-tuning on \fbow}
\label{tab:xlnet_fever}
\end{table}

\paragraph{Robustness of larger models}
We examine the performance of our method when applied to other PLMs and to larger networks by training BERT large and XLNET.
Fig~\ref{fig:base_large} shows the MNLI and HANS performance of those networks. Firstly, XLNet is noticeably more robust than BERT, compatible with its superior in-distribution performance~\citep{yang2019xlnet}.
Secondly, we observe that the large versions generalize on HANS significantly better than their base counterparts (e.g., 76.1\% vs 71.7\% for XLNet, 70.6\% vs 62.9\% for BERT), confirming that larger models seem more robust. Lastly, \xlnetlarge+\fbow shows a +7\% increase in performance, reaching 83.1\% on HANS with a maximum score of 86.8\% over three seeds.
We also show Table~\ref{tab:xlnet_qqp} and Table~\ref{tab:xlnet_fever}) similar findings on QQP and FEVER. 
For instance, \xlnetlarge+\fbow achieves 65.2\% on PAWS and 75.3\% on FEVER-Sym-v1.

\paragraph{Training on forgettables only} \citet{toneva2018empirical} showed that forgettable examples form the support of the training distribution. We follow their experimental setting and fine-tune BERT on the subset of forgettable examples only (i.e., without any fine-tuning on the whole dataset). Contrary to what was found in \citet{toneva2018empirical}, we observe in Table~\ref{tab:forg_tuning_mnli} that the performance obtained by training only on forgettable examples is poor compared to random subsets of the same sizes; MNLI accuracy is only 37.9\% for \fbert compared to 74.7\% for a random subset with the same size. The HANS accuracy is also poor. These results suggest an intrinsic difficulty in \fbert that makes it hard for \bertbase to generalize from it. However, as we showed previously, when starting from an already trained model, forgettables increase the out-of-distribution performance.

\begin{figure}[t]
\centering
\includegraphics[width=0.4\textwidth]{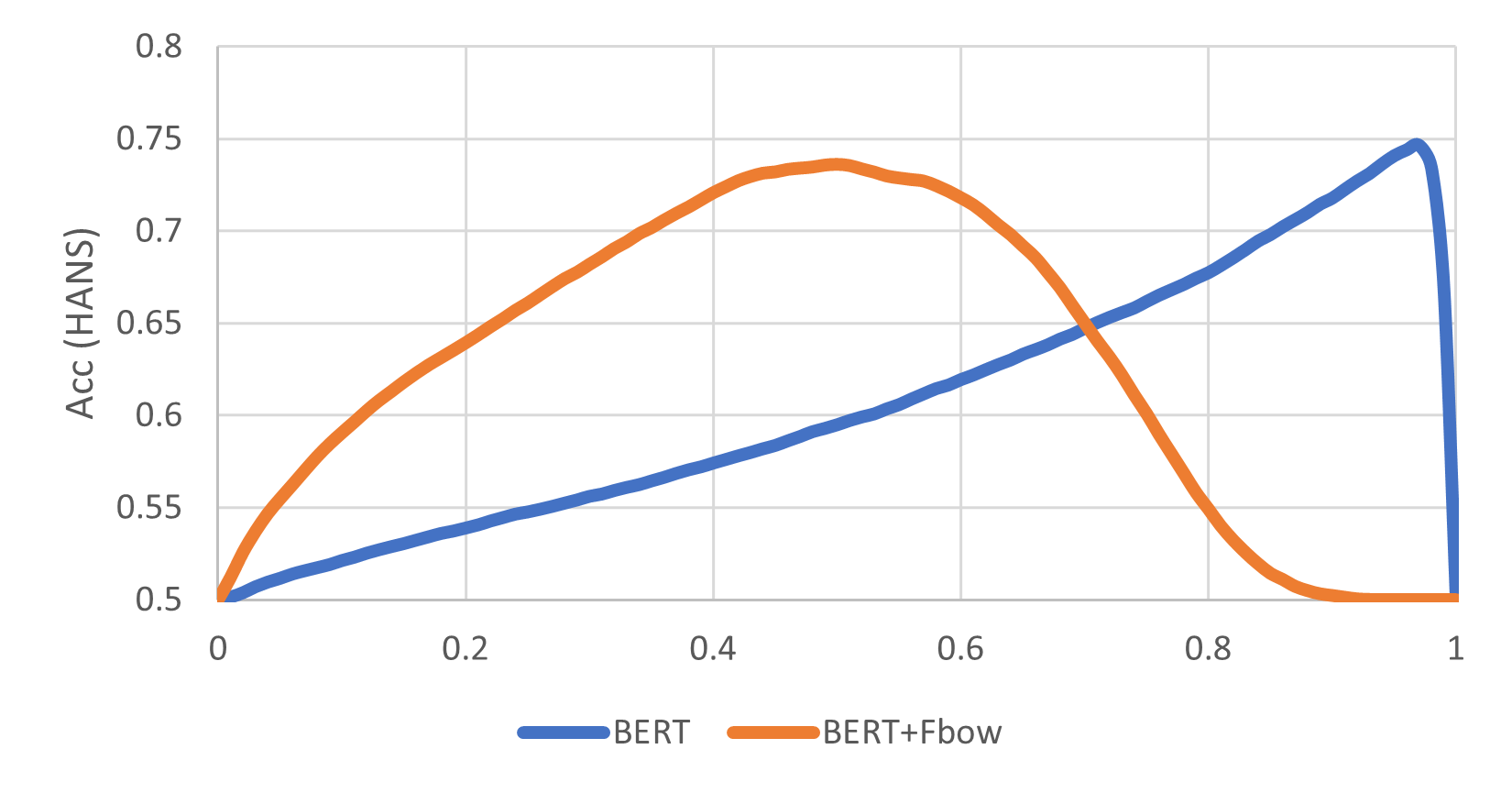}
\caption{HANS accuracy vs classification threshold used to predict entailment/non-entailment. The base BERT model is overconfident in the entailment class while after fine-tuning on forgettables, we can improve model calibration.}
\label{fig:threshold_hans}
\end{figure}

\paragraph{Calibration of models}
\label{sec:hans_callibration}
We look into the confidence of entailment when
\bertbase and \bertbase + \fbow trained on MNLI are applied to HANS.
In Fig~\ref{fig:threshold_hans}, we show that \bertbase can discriminate HANS entailments from non-entailments but with
a very large classification threshold. 
Fine-tuning on forgettables recalibrates the classification
threshold on HANS and makes 0.5 as the optimum value.

\paragraph{Other diagnostic evaluations}
Fine-tuning on the forgettable examples of simple biased models improves robustness in the three challenging benchmarks HANS, FEVER-Symmetric and PAWS. 
We additionally evaluate the trained models listed in Table~\ref{tab:maintable} on 
Stress tests~\citep{naik2018stress}, adversarial NLI~\citep{nie2019adversarial} and MNLI-matched-hard~\citep{gururangan2018annotation}. 
For these test sets, we do not observe improvements when evaluating the robust model using \fbow. 
We posit that specific biased models might be needed in some of these cases. As a validation, for MNLI-matched-hard, we design a BiLSTM model that only takes the hypothesis as input, and apply our method using the forgettables of that model to fine-tune \bertbase. We observe an increase in performance from 76.5\% to 78.0\% (averaged across five seeds).
These results suggest that the forgettable examples of simple biased models like BoW or Bi-LSTM capture the more informative heuristics like word-overlap well. However, for less informative heuristics like hypothesis-only features, a heuristic-designed biased model is a better choice since its forgettables likely violate the specific heuristic.

\section{Related Work}

A growing body of literature recently focused on \textbf{out-of-distribution generalization}, showing that it is far from being attained, even in seemingly simple cases~\citep{geirhos2018imagenet,jia2017adversarial,dasgupta2018evaluating}. In particular, and in contrast with what ~\citet{mitchell2018extrapolation} recommend, NLP models do not seem to ``embody the symmetries that allow the same meaning be expressed within multiple grammatical structures''. Supervised models seem to exhibit poor systematic generalization capabilities \citep{loula2018rearranging,lake2017generalization,baan2019realization,hupkes2018learning} thus seemingly lacking~\emph{compositional} behavior~\citep{montague1970universal}. While this might seem at odds with the common belief that high-level semantic representations of the input data are formed \citep{bengio2009learning}, the reliance on highly predictive but brittle features is not confined to NLU tasks. It is also a perceived shortcoming of image classification models \citep{geirhos2018imagenet,brendel19}.
To test systematically if machine learning models generalize beyond their training distribution, several challenging datasets have been introduced in NLP and other ML applications~\citep{journals/cmig/Kalpathy-CramerHDABM15,peng2018moment,clark2019dont}.
Those test sets are made automatically from designed grammars \citep{linzen2019right} and/or by human annotators~\citep{zhang-etal-2019-paws,schuster2019towards}.

\paragraph{Dataset re-sampling and weighting} These techniques have been studied in order to solve class imbalance problem~\citep{chawla2002smote} or covariate shift~\citep{sugiyama2007covariate}, notably by importance weighted empirical risk minimization. 
In NLP, \citet{clark2019dont,mahabadi2019simple,he2019unlearn,utama2020mind} give evidence of the effectiveness of re-weighting training examples to increase robustness. They generally assume~\emph{a priori} knowledge of the heuristics present in the dataset and up/down-weight examples concerning those heuristics. AFLITE~\citep{Sakaguchi2019winogrande} is an algorithmic method  for bias removal in datasets without relying on prior knowledge about datasets. It filters out examples with a high average predictability score, relating them to points with biases or spurious correlations. 
\citet{bras2020adversarial} adopt AFLITE to 
build more robust models. This method harms the in-distribution performance significantly. 
Comparatively, we aim to increase robustness while maintaining
in-distribution performance, so we do not filter out easy examples in our approach.
Our work also relates to distributionally robust optimization 
~\citep{duchi2018distrobust,hu2018} and the more recent group-DRO~\citep{sagawa20groupdrop}, which does not assume access to target data and optimizes the worst-case performance under an unknown, bounded distribution shift. 
Recently,~\citet{swayamdipta2020dataset} introduced a two-dimensional criterion to identify hard and easy examples. They consider both the confidence of an example (the average of its loss during training epochs) and the variability (those for which the loss has high variance) and show that ambiguous examples (high-variance and high-confidence) can enhance OOD accuracy. Easy examples (low-variance and high-confidence) can instead help model optimization. Although example forgetting is a coarser measure of variance of the loss, their results align with our findings: up-weighting hard/ambiguous examples enhance OOD generalization, but only training on those can harm optimization. 

\paragraph{Curriculum Learning} Dataset sampling is related to curriculum learning, where training proceeds along with a curriculum of samples with increasing difficulty~\citep{bengio2009curriculum}. \citet{Kumar10,Zhao2015,Fan2017,conf/icml/KatharopoulosF18,screenerNet,jiang18mentor} have shown the concept can be quite successful in a variety of areas.
Our robustifying method is related to this concept. However, our models are
first trained using i.i.d samples from the whole dataset and then fine-tuned on 
more difficult cases, i.e., the minorities.

\paragraph{Spurious correlations in NLU datasets}
like MNLI or FEVER are the subjects of many works. They include
(i) the presence of specific words in the hypothesis or claim, for example, negation words like ``not'' are correlated with the contradiction label in entailment tasks \citep{naik2018stress,gururangan2018annotation}, 
or bigrams like ``did not'' with the refute label, in fact, verification~\citep{schuster2019towards};
(ii) syntactic heuristics, like word-overlap between premise and hypothesis; and
(iii) sentence length \citep{gururangan2018annotation}, and its correlation with labels.
HANS \citep{linzen2019right} and PAWS \citep{zhang-etal-2019-paws} (which we evaluate on) generate plausible high word-overlap examples for both positive and negative classes.
\citet{glockner_acl18} build a new test example by
simple lexical inference rules and show the brittleness of models on this out-of-distribution dataset. 
They also show that having supporting examples in training data is key to predict a test example correctly.

\citet{feldman20longtail} show that when datasets are long-tailed, rare and atypical instances make up a significant fraction of the data distribution and \textbf{memorizing} them leads to better in-domain generalization. They find those rare and atypical examples using influence estimation.
We instead study forgettable examples and their impact on out-of-distribution generalization. An interesting experiment would be to mine minority examples by influence estimation and compare with  forgettable examples.


\section{Conclusion}
We introduced a novel approach, based on example forgetting, to extract minority examples and build more robust models systematically. Via example forgetting, we built a set of minority examples on which a pre-trained model is fine-tuned. We evaluated our method on large-scale models such as BERT and XLNet and showed a consistent improvement in robustness on three challenging test sets.
We also showed that the larger versions obtain higher out-of-distribution performance than the base ones but still benefit from our method. 

\section*{Acknowledgement}
We thank Rabeeh Karimi Mahabadi, Yoshua Bengio and all the anonymous reviewers for their insightful comments.

\bibliography{bibliography}
\bibliographystyle{acl_natbib}

\clearpage

\appendix

\section{Details of biased models (BoW and BiLSTM)}
\label{sec:biased_models}
Both models are Siamese networks, with similar input representations and classification layers.
For the input layer, we lower case and tokenize the inputs into words and initialize their representations with Glove, a 300-dimensional pretrained embedding~\citep{pennington2014glove}.
For the classification task, from the premise and hypothesis vectors $p$ and $h$, we build the concatenated vector $s = [p, h, |p - h|, p \odot h]$ and pass it to a 2-layer feedforward network. 
To compute $p$ or $h$, the BoW model max-pools the bag of word embeddings,
while the BiLSTM model max-pools the top-layer hidden states of a 2-layer bidirectional LSTM. The hidden size of the LSTMs is set to 200. Overall, BoW and BiLSTM contain 560K and 2M parameters, respectively.

\section{Hyperparameters and training time}
\label{sec:hyperparameters_trainingtime}
We use a learning rate of 5e-5 for MNLI and QQP when training
the PLMs on the full training and the learning rate of 1e-5 when fine-tuning on forgettables. For FEVER, we use 2e-5 and 5e-6 for the full training and the fine-tuning on forgettables, respectively.

With a 4x Tesla P100 GPU machine and batch-size 256 per GPU, one epoch of training on the full train set takes around 4-6 minutes for BOW and BiLSTM models in all of the three training tasks.

For \bertbase, with batch-size 32 per GPU, one epoch of training on the full train set takes around 30 / 20 / 30 minutes (per task).
The maximum input length after tokenization is set to 128 in all the experiments.

\subsection{Forgettables  and  word-overlap  in  MNLI}

\begin{table}[h]
\footnotesize
\setlength{\tabcolsep}{3pt}
\centering
\begin{tabular}{lccccccc}
\toprule
\textbf{Model} & \multicolumn{3}{c}{\textbf{Entailment}} & & \multicolumn{3}{c}{\textbf{Non-Entailment}} \\
& \emph{All}    & \emph{High}  & \emph{Low} & & \emph{All}     & \emph{High}    & \emph{Low} \\
\midrule
BERT & \textbf{84.0}   & \textbf{89.9}   & \textbf{76.0}  & & 84.9    & 85.5    & 84.6 \\

BERT + \fbow & 80.2  & 85.1 & 73.4 & &  \textbf{85.6} & 86.9 & \textbf{85.0}\\

BERT + \flstm & 79.9   & 85.2   & 72.4  & & \textbf{85.6}      & \textbf{87.4} & 84.8  \\
\bottomrule
\end{tabular}
\caption{Fine-grained accuracy results of \bertbase on the MNLI dev set split by word-overlap between hypothesis and premise.}
\label{tab:fine_mnli}   
\end{table}

In Table \ref{tab:fine_mnli}, we show the performance of our method on the MNLI dev set as a function of word-overlap, the main heuristic HANS was designed against. 
We split the evaluation set into High ($>$ mean) and Low ($<$ mean) word-overlap examples,
where word-overlap is measured using the Jaccard Index between hypothesis and premise.
We see in particular that entailment pairs with high word-overlap suffer from the fine-tuning on forgettables, while non-entailment improves
(we observe a similar trend for QQP; see App. \ref{sec:word_overlap_qqp}). This supports the observations in \ref{sec:forg-mionority} that the initial model relied on the spurious correlation of word-overlap and entailment to classify pairs and that by fine-tuning on forgettable examples, the performance on minorities increased.

\section{Forgettables  and  word-overlap  in  QQP}
\label{sec:word_overlap_qqp}
In Table \ref{tab:fine_qqp}, we show the performance of our method on the QQ evaluation set as a function of word-overlap, the main heuristic PAWS was designed against. We see in particular that paraphrase pairs with high word-overlap suffered from the fine-tuning, while non-paraphrase improved. This supports the intuition that the initial model relied on word-overlap to classify pairs as paraphrase, while forgettables help mitigate that phenomenon to some extent.

\begin{table*}[h]
\centering
\begin{tabular}{lccccccc}
    & \multicolumn{3}{c}{\textbf{Paraphrase}} & & \multicolumn{3}{c}{\textbf{Non-Paraphrase}} \\

                            Model    & \emph{All}    & \emph{High}   & \emph{Low}   & & \emph{All}     & \emph{High}    & \emph{Low}     \\
\toprule

BERT                       & \textbf{90.0} & \textbf{90.8} & \textbf{88.9} &  & 92.2 & 85.6 & 95.0 \\
BERT + \flstm & 85.2 & 84.9 & 85.8 &  & \textbf{93.0} & \textbf{87.3} & \textbf{95.4} \\   
BERT + \fbow    & 87.3 & 87.2 & 87.4 &  & 92.6 & 86.4 & 95.2 \\   
\bottomrule
\end{tabular}
\caption{Fine-grained accuracy results of BERT on QQP development set before and after fine-tuning on forgettables. 
We split the evaluation set into High ($>$ mean) and Low ($<$ mean) word-overlap examples,
where word-overlap is measured under the Jaccard Index between two sentences. Similar observations hold true in the case of MNLI.
}
\label{tab:fine_qqp}
\end{table*}

\end{document}